\documentclass[letterpaper, 10 pt, conference]{ieeeconf}  

\IEEEoverridecommandlockouts                              

\usepackage{graphicx} 
\usepackage{multirow}
\usepackage{afterpage}
\usepackage{subfig}
\usepackage{amsmath}
\usepackage{array}
\usepackage{tabularx}
\usepackage{booktabs}
\usepackage{url}
\usepackage[font=footnotesize, labelfont=footnotesize, textfont=footnotesize]{caption}
\usepackage{hyperref}
\usepackage{adjustbox}

\makeatletter
\newcommand{\setcaptype}[1]{\def\@captype{#1}}
\makeatother

\newsavebox{\tempbox}

\title{\LARGE \bf
Whole-Body Dynamic Throwing with Legged  Manipulators
}
\author{Humphrey Munn$^{1,2}$, Brendan Tidd$^{2}$, Peter B{\"o}hm$^{1,2}$, Marcus Gallagher$^{1}$, David Howard$^{2}$
\thanks{$^{1}$Humphrey Munn, Peter B{\"o}hm, and Marcus Gallagher are with the School of Electrical Engineering and Computer Science,
        University of Queensland, QLD. 4072, Australia
        {\{\tt\small h.munn,p.bohm,marcusg\}}{\tt\small@uq.edu.au}}
\thanks{$^{2}$Humphrey Munn, Brendan Tidd, Peter B{\"o}hm, and David Howard are with CSIRO Robotics, DATA61, CSIRO, Australia.
{\tt\small \{humphrey.munn,brendan.tidd,peter.bohm,\newline david.howard\}}
        {\tt\small@data61.csiro.au}}}%

\begin{document}

\maketitle
\savebox{\tempbox}{\begin{minipage}{\textwidth}
\centering
    \vspace{-0.6cm}
    \includegraphics[width=1\linewidth]{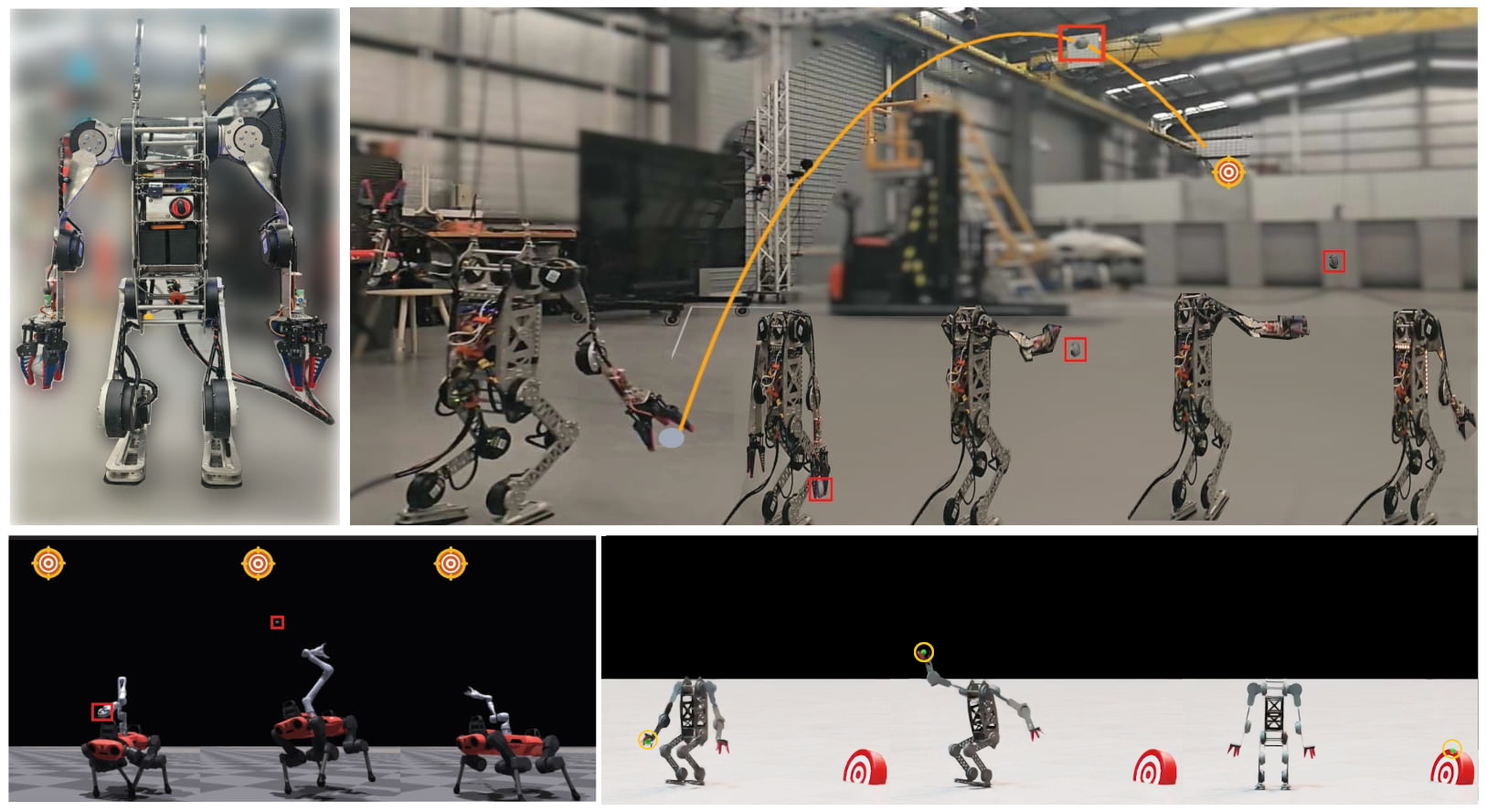}

    \captionof{figure}{\footnotesize Our robot throwing policies demonstrated on real hardware (top) and in simulation (bottom) showing complex full-body movements that improve throwing accuracy and distance, while maintaining a stable configuration. The top figure shows our humanoid setup and a long-range throw deployed sim2real and the real ball trajectory overlayed and highlighted in orange measured with motion capture. }
    \label{fig:heroshot}
    \vspace{-0.6cm}
\end{minipage}}
\begin{figure}[t]
\rlap{\usebox\tempbox}
\end{figure}
\afterpage{\begin{figure}[t]
\rule{0pt}{\dimexpr \ht\tempbox+\dp\tempbox}

\end{figure}}

\begin{abstract}

Throwing with a legged robot involves precise coordination of object manipulation and locomotion --- crucial for advanced real-world interactions. Most research focuses on either manipulation or locomotion, with minimal exploration of tasks requiring both. This work investigates leveraging all available motors (full-body) over arm-only throwing in legged manipulators. We frame the task as a deep reinforcement learning (RL) objective, optimising throwing accuracy towards any user-commanded target destination and the robot's stability. Evaluations on a humanoid and an armed quadruped in simulation show that full-body throwing improves range, accuracy, and stability by exploiting body momentum, counter-balancing, and full-body dynamics. We introduce an optimised adaptive curriculum to balance throwing accuracy and stability, along with a tailored RL environment setup for efficient learning in sparse-reward conditions. Unlike prior work, our approach generalises to targets in 3D space. We transfer our learned controllers from simulation to a real humanoid platform.         


\end{abstract}


\section{INTRODUCTION}
Throwing is a task all humans are familiar with: both because of its utility and because it is one of the most primitive ways to manipulate objects in our environment. Throwing an object towards a given position is often a faster method of transport than carrying it, which is crucial for efficiency purposes, e.g. in emergency scenarios. Furthermore, it allows deliverance of an object to places otherwise inaccessible, e.g. delivering a safety net or medical supplies to an area otherwise impossible to reach.


Complex dynamic manipulation, such as throwing, is inherently difficult due to the nonlinearity and uncertainty associated with high-speed motions~\cite{chen2024tossnet}.
The rapid nature of these movements requires accurate real-time sensor measurements, as well as actuators capable of delivering both high torque and precise control.

Although most existing research on robotic throwing employs mounted robot arms---using both analytical methods~\cite{gai2013motion,mori2009} and learning-based approaches~\cite{zeng2020tossingbot,ghadirzadehdeep,hu2010ball} to enhance stability---such setups narrow the range of achievable tasks and restrict the agility and throwing force attainable through full-body motion.

Whole-body throwing poses several challenges, especially due to the high degrees of freedom (DoF).
The robots tested exhibited high DoF with 13 (humanoid) to 19 (quadruped) actuators, resulting in a high-dimensional action space. Generalising body movement to a three-dimensional command space required high variability in joint trajectories for varied target locations. These layers of complexity do not lend themselves well to non-learning approaches, such as trajectory optimization or purely analytical methods~\cite{lynch1999dynamic}, which become computationally and methodologically infeasible to generalize across large action spaces.

A significant consideration is the trade-off between maximizing throwing capability and maintaining stability. Legged robots risk falling during high-performance maneuvers, which can lead to irrecoverable states or damage~\cite{green2016design}, especially when throwing over long distances or beyond the manipulator’s intended operating range.

In the context of DRL, throwing tasks pose a sparse reward challenge, as the primary reward (i.e., throw distance and accuracy) can only be evaluated once the projectile is released, and it is received only a single time. This means that the agent must execute a series of actions before experiencing delayed feedback, presenting a well-known difficulty for reinforcement learning algorithms~\cite{li2017deep}.

In this work, we present a method that improves both of these utilities by offering a solution capable of throwing accurately at any target in three dimensions, and by maximising the potential accuracy and range afforded by dynamic, full-body behaviours with legged manipulators.

We chose legged manipulators (or, loco-manipulators) as our platform of choice due to their novelty and affinity for this task, their general versatility, and for the associated complexities of a high DoF morphology for standard deep RL approaches. Legged robots offer advantages over wheeled robots by enabling traversal of rough terrains~\cite{hutter2017anymal, silva2012literature}, and with an added arm manipulator they can accomplish a wider set of possible tasks~\cite{fu2023deep}. 

We highlight three key aspects of our work.

\subsubsection{A novel formulation of full-body robotic throwing.}
We extend robotic throwing beyond traditional 2D target positions to a fully 3D target space, enabling throws to arbitrary locations. Unlike prior work limited to arm-mounted robots, our approach utilises full-body dynamics, leveraging reinforcement learning where trajectory optimisation and analytical methods fall short.

\subsubsection{Balancing throwing performance and stability.}
We introduce an adaptive curriculum with a sparse stability reward, dynamically optimising the trade-off between throw accuracy and balance. Our method uniquely optimises the curriculum thresholds to maximise both stability and performance, enabling high-velocity, accurate throws while maintaining balance. Our approach achieved some success when transferred to a real under-actuated humanoid robot although some limitations were present.

\subsubsection{Comprehensive performance analysis.}
We conduct an extensive evaluation across two robot morphologies, comparing full-body and arm-only throwing across different distances and throwing angles. Our results demonstrate superior accuracy, range, and stability on both a quadrupedal robot with an arm and a humanoid, highlighting the advantages of full-body motion for dynamic manipulation.

\section{RELATED WORK}

\subsection{Throwing Robots}
Numerous studies address the problem of throwing an object at a specified target using robotic systems. These approaches include analytical solutions with trajectory optimisation techniques~\cite{gai2013motion,mori2009,mason1993dynamic,senoo2008high,taylor2019optimal}, and learned policies~\cite{zeng2020tossingbot,ghadirzadehdeep,hu2010ball} derived from simulation or real-world data.

Most focus solely on an arm mechanism; notable exceptions include~\cite{abe2013dynamic}, where trajectory optimization is used to enable an armed quadruped to throw, and~\cite{liu2022solution}, which applies reinforcement learning to throwing with a wheeled base.

The throwing task has been demonstrated using constrained solutions that typically solve only for projectile release velocity, with other movement parameters and release positions held constant. While effective for objects that vary along a 2D plane, this restriction prevents targeting arbitrary points in 3D space. Furthermore, these constraints are not applicable for a full-body throw as they do not account for the additional dynamics, flexibility, and interaction provided by a full-body robot with a base that moves non-linearly.

In~\cite{hu2010ball}, a 99\% accuracy rate was achieved for throws aimed at a target 3m away from various angles by parameterizing the ball’s release speed with a cubic function of the target position and estimating the function coefficients from real-world data. Although this approach is successful for a single plane, it lacks adaptability for targets that differ in height and applies only to a robot arm, foregoing the flexibility and momentum benefits of a full-body system.

Similarly,~\cite{zeng2020tossingbot} solves for the projectile release velocity with an arm, and generalises throws into different bins spaced apart on a 2D plane with arbitrarily shaped objects. Another constrained solution is shown in~\cite{liu2022solution}, however, utilises a wheeled base to generate additional momentum for throws.

Trajectory optimisation (TO) with behaviour constraints is used in~\cite{abe2013dynamic} to enable an armed quadruped to throw a cinder block, utilising the body momentum to throw the block. This study highlights the advantages of the robot body and the challenges of throwing well while maintaining stability. However, the solution lacks generalisation to arbitrary targets in space or generalisability of non-linear trajectories (e.g. air resistance). Our method is able to replicate this behaviour without the limitations of TO.

Several recent solutions have utilised RL to successfully demonstrate the task of throwing a projectile at a target. An excavator with an arm is trained in~\cite{werner2024dynamic} using a simulation to throw accurately at a target that varies along the x-y plane. The reward formulation is a function of the minimum distance between the ball and the target, similar to our work.

Bi-manual setup trained with RL to throw between two arms is shown in~\cite{huang2023dynamic, lan2023dexcatch}. These works successfully transfer policies trained in a distributed simulation environment onto real hardware successfully. However, the task of passing between hands is done in close proximity which is very different to the goal of our work.

\subsection{Full-Body Learning}

Recent works have explored tasks involving a manipulator and a legged base, dubbed loco-manipulation, and associated challenges. Full-body policy that can be commanded to walk and use a manipulator for reaching or grasping is presented in~\cite{fu2023deep}. The robot body assists the manipulator by extending its range with a unified RL controller for the base and arm. A curriculum is used to minimise local minima by gradually mixing locomotion and manipulation advantage estimates in the PPO surrogate loss.

However, for our task, an early focus on stability leads to policies that are too conservative to explore dynamic motions like jumping to maximise throwing momentum. Therefore, we customise our curriculum design to avoid potential local minima arising specifically from the throwing task.

The problem and difficulty of local minima for a set of loco-manipulation tasks is tackled in~\cite{wang2024arm} by training RL policies with curriculum learning while successfully transferring policies to a real robot. A loco-manipulation policy for box picking with a humanoid is trained in~\cite{dao2024sim}. The RL framework decomposes the task and reward shaping to achieve effective results.
However, this decomposition reduces the flexibility of learnt policies and a decomposition for the throwing task is not obvious and dependent on the throwing angle.













\section{METHOD}





Our method learns a full-body joint-level controller that optimises throwing accuracy and stability while reducing the effects of local optima through a simple adaptive curriculum without imposing constraints on the agent’s behaviour. The model is trained in Isaac Lab ~\cite{makoviychuk2021isaac}, following the massively parallel PPO implementation from~\cite{rudin2022learning}. A ball is initialised in the robot's hand at the start of each run, and a throwing target in polar coordinates is provided as input to the policy. Throwing accuracy and stability are measured during each run and used to calculate rewards for the agent. For hardware validation, we transfer our model to a humanoid robot for system verification.

\subsection{Throwing Task Definition}
Let $T$ be a $N$ x $3$ matrix representing a trajectory of projectile positions relative to the robot frame $(x_t,y_t,z_t)$ in meters over $N$ time steps, where $N$ is the first instance where the ball reaches the ground-plane $z_t = 0$. The trajectory starts when $(x_0,y_0,z_0)$ is $\geq$ 25cm away from the origin of the end-effector to avoid potential finger collisions. Let A be the target position vector of the projectile (desired destination) $(x_A, y_A, z_A)$. Then, the throwing displacement error $E$ can be written as:
\begin{equation}
E = \min_{t \in \{1, 2, \ldots, N\}} \left\| \begin{bmatrix}
x_t \\
y_t \\
z_t
\end{bmatrix} - \begin{bmatrix}
x_A \\
y_A \\
z_A
\end{bmatrix} \right\|_2
\label{eq:displacement}
\end{equation}

The minimum Euclidean distance is taken to consider different trajectories that enter within the same distance of the target as equally desirable.

The target positions are parameterised with spherical coordinates $\tilde{\theta}, \phi,$ and $r$,  where $\tilde{\theta}$ is a normalised polar angle representing the cosine of the standard polar angle $\theta$, $\phi$ is the azimuthal angle and $r$ is the radius of the sphere at the target distance. These coordinates are then converted to Euclidean space.

Two separate throwing tasks are presented in this work, \textbf{distance throw} and \textbf{generalised throw}. For the distance throw task, $\tilde{\theta}$ and $\phi$ are set to 0. For the generalised throw task, $\tilde{\theta} \in [0,1], \phi \in [0, 2\pi]$, which encapsulates all possible targets above the ground plane at a distance of $r$. For both tasks, the radius parameter $r$ is sampled within a range, see Section~\ref{sec:cldesign}. For both tasks, targets are generated by uniformly sampling within these ranges.

\subsection{Reinforcement Learning Formulation}

Deep Reinforcement Learning is used to optimise a policy \(\pi\) by aiming to maximise the discounted cumulative return \(E_\pi \left[ \sum_{t=0}^{T-1} \gamma^t r_t \right]\), where \(r_t\) denotes the reward at time step \(t\), \(\gamma\) is the discount factor, and \(T\) represents the maximum length of an episode. For stability experiments, \(r_t\) aggregates throwing rewards and stability rewards, otherwise consists only of the throwing reward.

\subsubsection{Optimisation Objectives}


We constructed a minimal representation of the task reward without extensive reward shaping, enabling greater flexibility in the policy behaviour. 

The reward was calculated as \( r = \lambda_1 r_{\text{throwing}} + \lambda_2 r_{\text{stability}} + \lambda_3 r_{\text{roll}} \), with \( \lambda_1 \), \( \lambda_2 \), and \( \lambda_3 \) as the reward coefficients. The reward coefficients for the humanoid follow Table \ref{tab:ablation}, whereas the quadruped coefficients were manually tuned to $(\lambda_1,\lambda_2,\lambda_3) = (1.0, 0.1, 0.0)$. The auxiliary reward penalty $r_{\text{roll}}$ was required for stability in the humanoid as its under-actuated legs cannot perform abduction which is necessary to recover from significant rotation along the roll axis. 


The throwing reward component is: \( r_{\text{throwing}} = 1 - \min \left[\frac{E}{r}, 1\right] \), where \( E \) is from Equation~\ref{eq:displacement} and \( r \) is the target distance. This results in a linear reward scaling based on the normalised throw displacement, given as a sparse reward when the distance between the ball and end effector exceeds 25 cm. Rather than waiting for the ball to complete its trajectory in simulation, the reward is calculated using projectile motion equations, minimising the delay between reward and relevant policy actions, thus improving performance (see Table \ref{tab:ablation}). This also allows the policy to learn non-linear motion by rewarding throws based on a drag model (e.g., Newtonian drag). We verified similar results with the Newtonian drag model but used standard projectile equations for simplicity, as the trajectories were nearly identical between the two models.

The unscaled stability reward, \( r_{\text{stability}} \), is 1 if certain conditions are met; otherwise, it is 0. This reward is added to the overall reward \( r \) at the episode’s final step, maintaining performance similar to a non-sparse reward (see Table \ref{tab:ablation}). For \( r_{\text{stability}} > 0 \), the policy must satisfy several criteria to mitigate local minima early in training, including preventing collisions (except with the feet), keeping the robot’s base height within a specified range, and ensuring the ball is thrown.

Roll rotation outside the humanoid's safe margin of (-0.1, 0.1) radians was penalized with the reward component \( r_{\text{roll}} = \frac{e^{(1-\frac{|roll|}{0.1})} - 1}{1+e^{-10\cdot (|roll| - 0.3)}} \), ensuring \( r_{\text{roll}} \geq -1 \) with a small positive reward within the safe margin. This dense feedback was added at each time step to aid in maximising \( r_{\text{stability}} \).

\subsubsection{State and Action Representation}

For both robots, the state representation as input to the policy \(\pi\) includes linear velocity from the IMU, throwing target commands (adjusted polar angle $\tilde{\theta}$, azimuthal angle $\phi$, radius r), motor state information (relative joint rotation, joint velocity), and the previous action output by the policy. Additionally, the observation space is augmented with a binary ('ball released') observation, as well as the estimated throwing displacement the robot would receive if the ball was let go of at the current time-step. We found both of these observations to be beneficial: see Table \ref{tab:ablation}. As the latter observation is privileged information only available in simulation, it is gradually replaced in a curriculum with noise of equal distribution over 100 training iterations with minimal performance impact.   



For position control, actions are implemented through a simulation of DC motor dynamics provided by IsaacLab to facilitate sim2real transfer. An additional action controls ball release if it exceeds 1.   



\subsubsection{Neural Network Configuration and Learning Parameters}

A recurrent symmetric Actor-Critic network architecture was trained with DPPO.
Hyper-parameters differing from \cite{rudin2022learning} were a network architecture of size $[256, 128, 64]$, clipping parameter 0.15, environment steps = 26, ELU activation, entropy = 0, optimiser = AdamW optimiser, $\lambda$ = 0.93, mini-batches = 6, initial action noise = 0.5.  All hyperparameters were the same for all experiments. We found with a larger neural network, a GRU backbone implemented following 
\cite{bohm2023feature} was beneficial, but was not used after the remaining hyperparameters were optimised (see Table \ref{tab:ablation}).

\subsection{Curriculum Design}
\label{sec:cldesign}
The proposed curriculum offered several benefits including balancing the trade-off between stability and throwing accuracy, improving convergence time, and improving throwing performance by helping the policy navigate out of local minima. Two curricula are proposed for each robot morphology as each exhibit distinct locally optimal behaviours. For example, the quadruped base has a greater degree of control with four legs so policies risk learning to throw primarily with body momentum rather than the arm.

The throwing and stability performance thresholds for the adaptive difficulty curriculum were learned via an optimisation process (see Table \ref{tab:ablation}), which significantly improved policy performance. This contrasts existing work using manually tuned curriculum thresholds (e.g., \cite{rudin2022learning,okamoto2021reinforcement,howard2022assessing}). When the criteria are met at a particular learning iteration, the difficulty advances 1\% of the distance range until reaching the maximum distance. If the policy did not reach the maximum distance, the maximum distance was gradually increased to the maximum distance for the final 100 training iterations.

The \textbf{general} throwing task (command: $\tilde{\theta},\phi,r$) curriculum was identical for each robot compared with the \textbf{distance} throw (command: $r$) curriculum except for the range of uniformly sampled target distances, which was widened linearly from \([1\,\text{m}, 3\,\text{m}]\) to \([1\,\text{m}, \text{max\_dist}]\).

\subsubsection{Quadruped Curriculum (Distance Throw)}
For the \textbf{distance} throw, the target position was fixed to (\( \text{max dist} \)) for all training iterations rather than using the curriculum schedule. This encouraged the maximum use of momentum by the high DoF arm. Momentum generation in the arm was further promoted by gradually introducing body actions in increments of 2\%, from 0\% to 100\%. This value was multiplied by the policy actions on all non-arm motors for full-body experiments.

Additionally, the wait time required to receive the stability reward after ball release was increased incrementally from 0 to 2 seconds, with a step size of 0.02 seconds. This enabled both the stability objective to increase in difficulty gradually, and the throwing performance to converge faster by initially reducing the episode time, ending as soon as the ball is released with a 0-second wait. Each of the above curriculum steps was executed based on the adaptive curriculum criteria.

\subsubsection{Humanoid Curriculum (Distance Throw)}

For the \textbf{distance} task, the humanoid's target distance was incrementally increased from 4 metres to a specified maximum distance according to the adaptive schedule. Starting with shorter target distances allowed the policy to learn a balance between stability and throwing performance before increasing the task difficulty with targets further away. 

\section{EVALUATION}
\begin{figure}[!b]
    \vspace{-0.4cm}
    \centering
    \includegraphics[width=119pt]{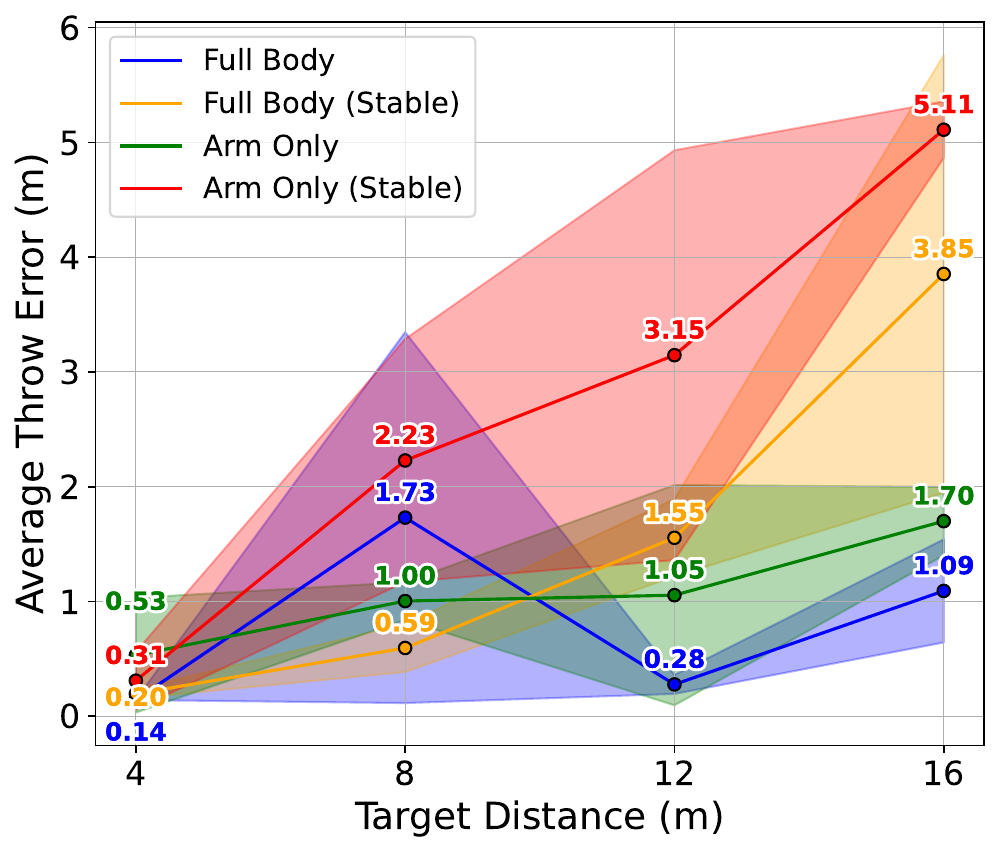}
    \includegraphics[width=123pt]{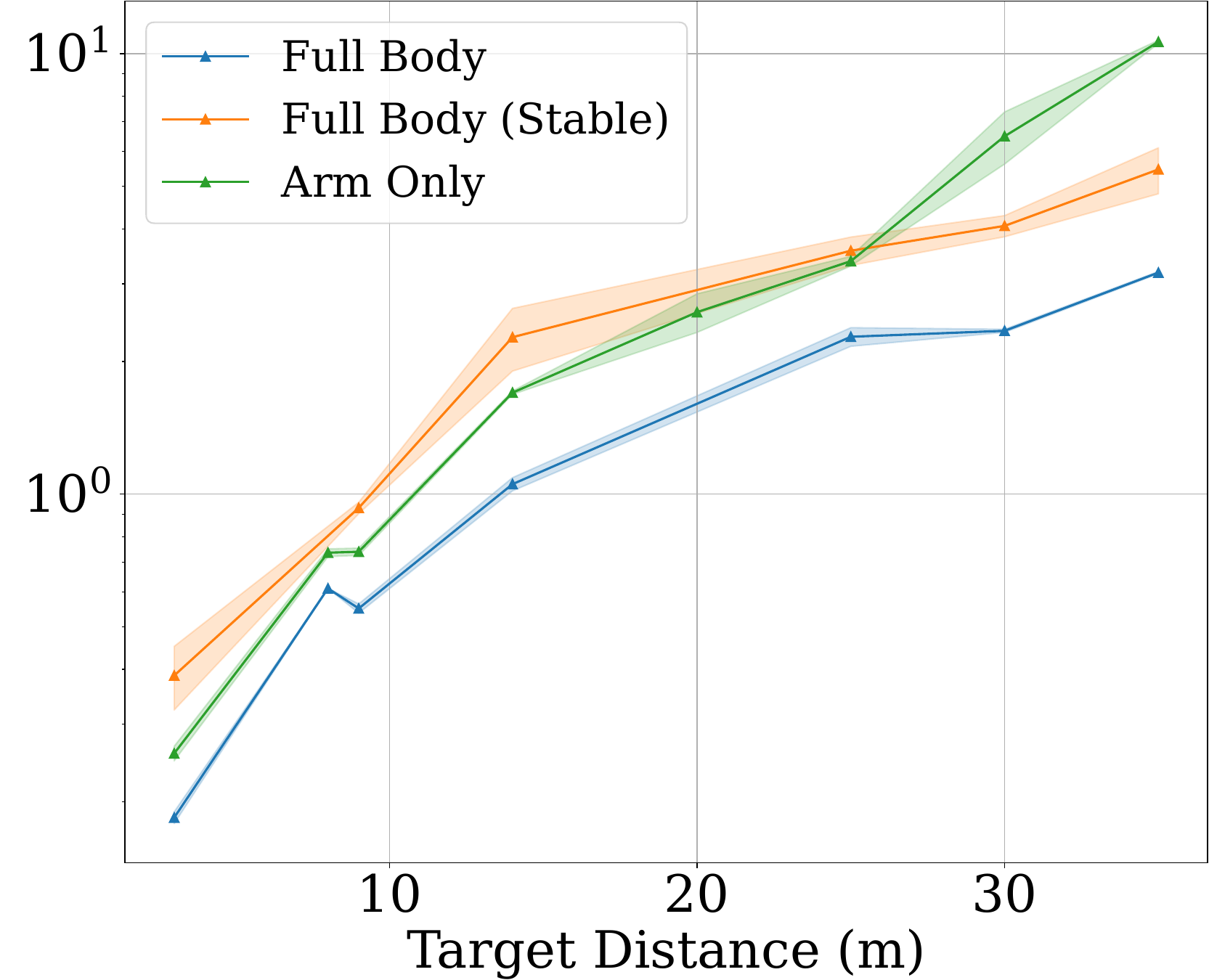}
    \caption{\footnotesize Throwing error (metres) between robot throw and target for the humanoid (left) and the quadruped (right) in simulation experiments. Shaded region represents $\pm$ 1 standard deviation over 5 runs. PyTorch random seed was randomised only for humanoid experiments.}
    \label{fig:dist_error}
    \vspace{-0.2cm}
\end{figure}

\begin{figure}[t]
\centering
   \vspace{1.25cm}
   \includegraphics[width=\columnwidth,trim={0pt 0pt 0pt 30pt}]{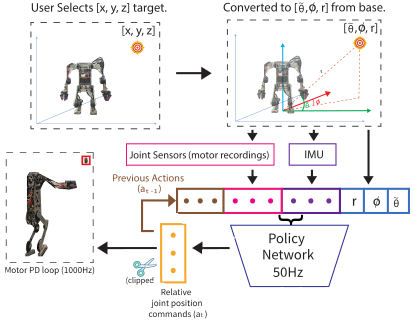}
   \caption{ Policy Evaluation and Real-world deployment architecture.}
   \label{fig:systemdiagram}
   \vspace{-0.5cm}
\end{figure}

We evaluated our policies in terms of throwing accuracy and percentage of stable runs. We test the generalisability of our method to different throwing angles and distances ( \textbf{general} task), as well as our method's ability to throw at fixed targets at various distances (\textbf{distance} task). Figure~\ref{fig:systemdiagram} illustrates the process of executing each policy.

\subsection{Experimental Setup}
We conducted simulation experiments with two robots, a humanoid and a quadruped, using physics parameters from~\cite{rudin2022learning} across 4092 parallel environments. To accommodate the more complex collisions resulting from ball contact with the gripper, we artificially fixed the ball's position to the hand until the policy issued a release command. We added a 100-ms delay after the command until the ball was released to account for the time it took for the real hand to open. Furthermore, we clipped all policy actions to within the joint operating range of the respective robots. Real-world experiments were conducted solely with the humanoid robot.


The ANYbotics ANYmal C quadruped was equipped with a Kinova Gen3 six-degree-of-freedom arm and a three-finger gripper attached at the base link. The quadruped was initialised in its default stance, and the arm was set to a home position with its fingers closed. In simulation, the base and arm employed proportional-derivative (PD) gains of 80/1 and 20/1, respectively. These settings prevented the base from collapsing under the arm’s weight and ensured the arm could achieve sufficient throwing velocity without causing instability or unrealistic behaviour.

The humanoid robot, named ``Chuck'', is an in-house design with 12 DoF and two soft, three-finger FinRay grippers, each hand actuated by a single servo motor.
The primary DoFs reside in the ankles, knees, hips, shoulders (two axes each), and elbows on both sides of the robot. For both simulation and real-world experiments, the base and arm employed PD gains of 40/1 and 10/1, respectively. Each joint is driven by a variant of the MIT Cheetah actuator from~\cite{bledt2018cheetah}, which includes a 6:1 planetary gear reduction and an onboard PD control loop operating at 1kHz.

For simulation, we utilise STL collision meshes and measured each link’s weight to build a URDF model of the humanoid robot, while the quadruped used its default publicly available URDF in home position.
In both simulation and real-world experiments, Chuck was initialised in a stable position with arms lowered, knees fully bent, fingers closed, and the torso upright (see Figure~\ref{fig:heroshot}, top left).

All configurations used relative joint-position motor commands at a frequency of 50Hz. 
The policies were trained on flat ground, with domain randomisation employed to assist in sim2real transfer and policy exploration, following the parameters from Table \ref{tab:domain_rand}. 

\begin{table}[h]
\centering
\renewcommand{\arraystretch}{0.8}
\begin{adjustbox}{max width=\columnwidth}
    \begin{tabular}{l c}
        \toprule
        \textbf{Parameter} & \textbf{Range} \\
        \midrule
        Right arm joints init. & $\pm 0.3$ rad \\
        Other joints init. & $\pm 0.025$ rad \\
        Root lin. vel. noise & $\mathcal{U}(-0.05, 0.05)$ \\
        Joint pos. noise & $\mathcal{U}(-0.01, 0.01)$ \\
        Joint vel. noise & $\mathcal{U}(-0.05, 0.05)$ \\
        Body mass variation & $\pm 0.5$ kg \\
        \bottomrule
    \end{tabular}
\end{adjustbox}
\caption{Domain Randomisation Parameters}
\label{tab:domain_rand}
\vspace{-0.3cm}
\end{table}

To enhance stability, policy actions were zeroed 250ms after the ball was released, allowing the robot to revert to its home state. This delay provided enough flexibility for repositioning the robot into a stable posture while reducing the complexity of the stability task.

\subsection{Simulation Experiments Results}

\begin{table}[t!]
    \setlength{\tabcolsep}{6pt}  
    \renewcommand{\arraystretch}{0.8}  
    \centering
    \scriptsize
    \begin{adjustbox}{max width=\columnwidth, min width=\columnwidth}
    \begin{tabular}{lccc}
        \toprule
        \textbf{Parameter} & \shortstack{\textbf{Importance} \\ (\%)} & \shortstack{\textbf{Top Trial} \\ \textbf{Value}} & \shortstack{\textbf{Mean/Mode} \\ \textbf{(Top 10 Trials)}} \\ 
        \midrule
        Stability Reward \%         & 28.65  & 0.02  & 0.04  \\ 
        Stability Threshold         & 13.95  & 0.22  & 0.47  \\ 
        Accuracy Threshold          & 12.49  & 0.51  & 0.41  \\ 
        GRU                         & 9.87   & False & False \\ 
        Desired KL                & 8.78   & 0.02  & 0.02  \\ 
        Roll Reward \%              & 8.06   & 0.17  & 0.12  \\ 
        Reward Scale                & 7.94   & 2.54  & 2.72  \\ 
        Value Loss Coefficient      & 6.20   & 0.98  & 0.84  \\ 
        Global Foot Pitch State     & 1.22   & False & False \\ 
        Body Roll State             & 1.18   & False & False \\ 
        Estimate Displacement State & 1.07   & True  & True  \\ 
        Ball Released State         & 0.59   & True  & True  \\ 
        \bottomrule
    \end{tabular}
    \end{adjustbox}
    \vspace{10pt}  
    \begin{adjustbox}{max width=\columnwidth, min width=\columnwidth}
    \scriptsize
    \begin{tabular}{lcc}
    \toprule
        \textbf{Parameter} & \shortstack{\textbf{Average Throwing} \\ \textbf{Accuracy (\%) Improvement}} & \shortstack{\textbf{Average Stability} \\ \textbf{Improvement (\%)}} \\
    \midrule
    Projectile Motion Reward Prediction & +55.2 & +84.2 \\
    Sparse Stability Reward & +1.0 & -3.7 \\
    \bottomrule
    \end{tabular}
    \end{adjustbox}
    \vspace{-0.1cm}
    \caption{Ablation study results in simulation for the \textbf{general} humanoid policies. Top: Results across 700 trials using the Optuna optimiser and default parameter importance metric. Bottom: Results averaged over 5 randomised trials.}
    \label{tab:ablation}
    \vspace{-0.1cm}
\end{table}

For the \textbf{general} throwing task, we sampled distance commands from $U \left(1, 5 \right)\, \text{metres}$ and $U \left(3, 8 \right)\, \text{metres}$ for the humanoid and quadruped robots respectively. This range was within the throwing power of both arms, enabling accuracy comparisons between policies rather than focusing on maximum throwing power.

Figure~\ref{fig:dist_error} demonstrates the accuracy of the throws for each robot and the body configuration for the \textbf{distance} throw task, where targets were placed at various distances in front of the robot on the ground plane.
The relationship between throwing accuracy and target distance shows some exponential growth, particularly for the arm-only experiments. This is because slight differences in the compound error of the projectile release velocity over larger distances are known phenomena in throwing sports~\cite{reina2018throwing, wilson2016using}.

\begin{table}[!h]
\vspace{0.15cm}
\vspace{-0.15cm}
\centering
\scriptsize
\begin{adjustbox}{max width=\columnwidth, min width=\columnwidth}
\begin{tabular}{l|cc}
\toprule
\multirow{2}{*}{\textbf{Category}} & \multicolumn{2}{c}{\textbf{Mean (m) $\pm$ Std (m)}} \\
 & \textbf{Humanoid} & \textbf{Quadruped} \\
\midrule
Arm Only & na & 0.740 $\pm$ 0.014 \\
Arm Only (Stable) & 1.361 $\pm$ 0.342 & na \\
Full body & na & \textbf{0.550 $\pm$ 0.013} \\
Full Body (Stable) & \textbf{0.914 $\pm$ 0.143} & 1.092 $\pm$ 0.066 \\
\bottomrule
\end{tabular}
\end{adjustbox}
\caption{\footnotesize Error for \textbf{General} Policies in Simulation (5 trials)}
\label{tab:throw_error_comparison}
\vspace{-0.2cm}
\end{table}

Full-body policies, aside from one trial, were consistently more accurate than arm-only policies over all distances. Visually, when the target is beyond the arm's maximum range, the behaviour of the full-body controller exhibits novel behaviour, such as utilising stored elastic energy from the body to generate additional velocity \cite{roach2014upper}, counterbalancing with the left arm on the humanoid, and stepping with the feet for stability. Arm-only policies also generated momentum by utilising a winding-up motion before extending for a high-velocity throw, most evident on the quadruped with 6DoF arm.

\begin{table}[!h]
\centering
\begin{adjustbox}{max width=\columnwidth, min width=\columnwidth}
\begin{tabular}{l|cc}
\toprule
\multirow{2}{*}{\textbf{Category}} & \multicolumn{2}{c}{\textbf{Mean (\%) $\pm$ Std (\%)}} \\
 & \textbf{Humanoid} & \textbf{Quadruped} \\
\midrule
General Throw & 84.2 $\pm$ 5.89 & 81.8 $\pm$ 3.4 \\
Distance Throw (4m;3m) & 98.1 $\pm$ 1.7 & 91.1 $\pm$ 1.7 \\
Distance Throw (8m;14m) & 29.0 $\pm$ 38.1 & 88.6 $\pm$ 4.7 \\
Distance Throw (12m;25m) & 48.3 $\pm$ 48.00 & 66.1 $\pm$ 1.5 \\
Distance Throw (16m;35m) & 79.1 $\pm$ 18.9 & 10.0 $\pm$ 11.7 \\
\bottomrule
\end{tabular}
\end{adjustbox}
\caption{\footnotesize Stability of Throwing Policies in Simulation (5 trials)}
\label{tab:stability_error_comparison}
\vspace{-0.3cm}
\end{table}

\begin{figure}[!h]
    \centering
    \includegraphics[height=100pt, trim={20pt 90pt 194pt 175pt}, clip]{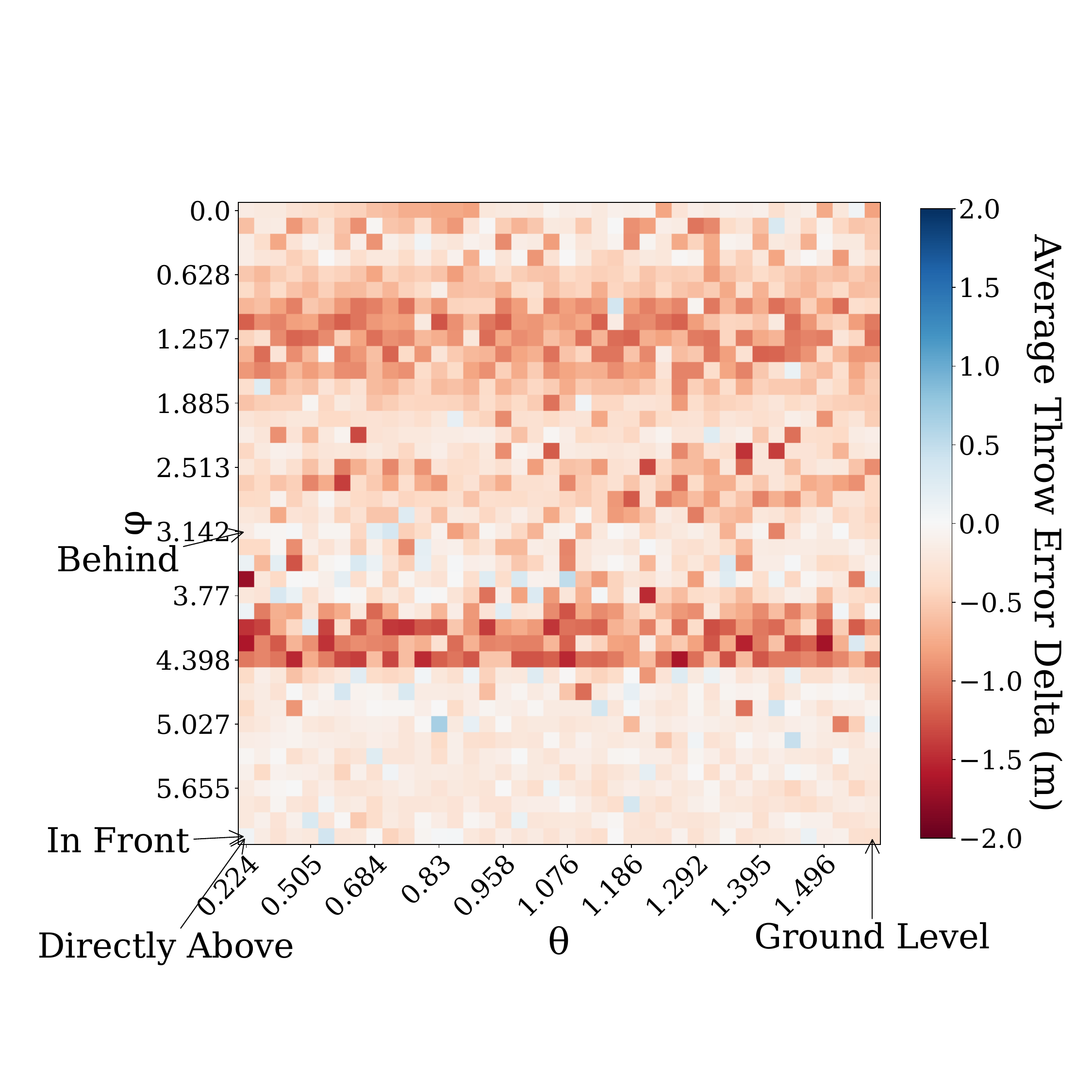}
    \includegraphics[height=100pt, trim={20pt 90pt 28pt 175pt}, clip]{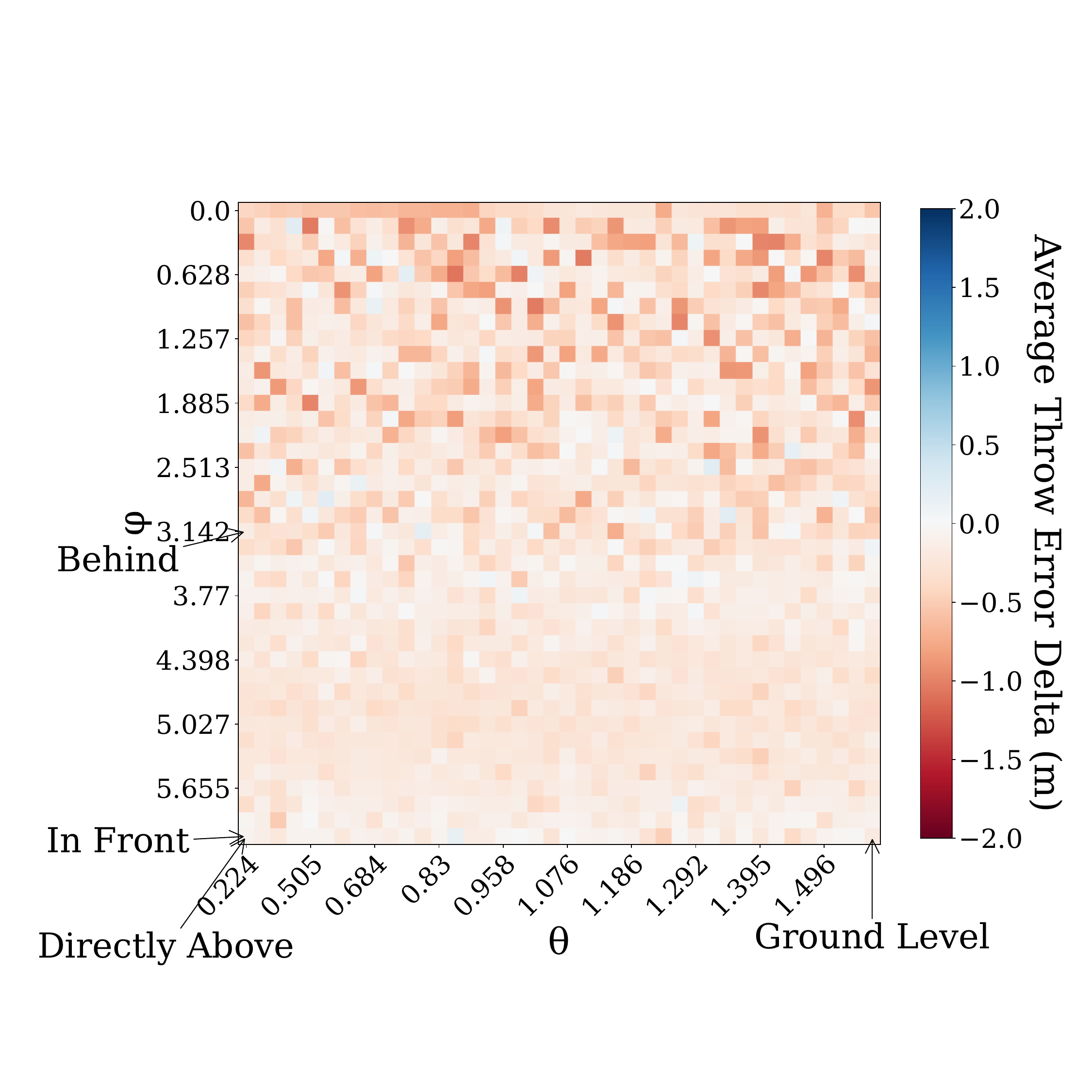}
    \caption{\footnotesize Throwing error difference (\emph{full-body policy} $-$ \emph{arm-only policy}) from the target (metres) for the humanoid (left) and quadruped (right), at a distance of 5 and 8 metres respectively. Error difference displayed over all discretised throwing angles ($\phi$,$\theta$). Averaged over 3 trials per angle. Red indicates higher performance from full-body.}
    \label{fig:polar_diff}
    \vspace{-0.3cm}
\end{figure}

We measured the consistency of the full-body policies across all discretised throwing angles ($\phi$ and $\theta$) in simulation, demonstrated in Figure \ref{fig:polar_diff}. The full-body policies were consistently more accurate than arm-only policies, as shown by more than 99\% of throwing angles having greater accuracy for the full-body policy than arm-only. In particular, the humanoid full-body policy shows the greatest improvement over the arm-only policies for throwing angles directly to the left or right of the humanoids's heading direction. 


We present in Table \ref{tab:stability_error_comparison} the average stability for each configuration. Our stability metric requires the base height to never drop below a safe threshold,
no collisions to occur between the links (excluding feet) and the ground, and for the policy to have released the ball to ensure training stability. We measured stability including domain randomisation (Table \ref{tab:domain_rand}) and action sampling for robustness. The large variability in stability across the humanoid trials results from the randomisation of the learning algorithm seed. 

Finally, we did an ablation of various system components as shown in Table \ref{tab:ablation} (top). Due to the large number of parameters, we took a black-box optimisation approach to obtain parameters using asynchronous tree-structured parzen estimation (TPE) optimisation. TPE was more efficient and led to greater results compared with manual tuning or random search. The objective was the sum of throwing accuracy and clipped stability rate: $r_{\text{throwing}} + \min\left[r_{\text{stability}}, 75\%\right]$. Of particular importance were the parameters of the adaptive curriculum schedule and the reward coefficients.

\subsection{Real-world Experiments Results}
Finally, we evaluated our method by transferring a learned policy onto the humanoid. Quantitative results are presented in Table \ref{tab:realworld} for a full-body, stable throw at a target distance of 4 metres. Ball trajectories were tracked with motion capture. Out of 5 real-world trials, our policy achieved an average error of 80cm while remaining stable throughout throws. Additionally, we tested the robot with a 1.25kg payload, which remained stable and achieved comparable accuracy. We encountered difficulties with sensor misalignment, particularly for policies exhibiting rapid movements. This is an area for future work. Some trajectories generated in simulation for a humanoid policy trained to throw at a distance of 12 meters were successfully deployed in an open-loop setting, maintaining stability while achieving a throw distance of 13.8 meters -- 4.95 meters further than the maximum distance achieved with a stable arm-only policy.

\begin{table}[h]
\centering
\renewcommand{\arraystretch}{0.8} 
\begin{adjustbox}{max width=\columnwidth}
    \begin{tabular}{lcc}
        \toprule
        & Throwing Error from Target & Stability \\
        \midrule
        Best Run  & 6 cm  & Stable (No Fall) \\
        Average   & 80 cm  & Stable (No Fall) \\
        \bottomrule
    \end{tabular}
\end{adjustbox}
\caption{Real-world throwing error from the humanoid policy at a target of 4 metres, averaged over 5 Trials.}
\label{tab:realworld}
\vspace{-0.3cm}
\end{table}

\section{CONCLUSION}

We presented a method for throwing a projectile to any point in three-dimensional space using full-body control, with best-performing policies achieving an average throwing error of 73cm from targets anywhere within a 5-metre radius of the humanoid robot, and comparable results in simulation for the quadruped. We showed that the body offers significant improvements over arm-only methods despite stability constraints, aided by our novel adaptive curriculum and environment setup. We demonstrated the approach successfully on hardware and evaluated our deep RL approach in simulation with both an armed quadruped and a humanoid.





\addtolength{\textheight}{-5cm}   

\bibliography{references}

\end{document}